\documentclass{article} 
\usepackage{iclr2020_conference,times}

\usepackage{amsmath,amsfonts,bm}

\def\eqref#1{equation~\ref{#1}}

\def\1{\bm{1}}

\DeclareMathAlphabet{\mathsfit}{\encodingdefault}{\sfdefault}{m}{sl}
\SetMathAlphabet{\mathsfit}{bold}{\encodingdefault}{\sfdefault}{bx}{n}

\usepackage{hyperref}
\usepackage{url}
\usepackage[ruled,vlined]{algorithm2e}
\include{pythonlisting}
\usepackage{graphicx}
\iclrfinalcopy

\title{Multi-Task Learning in Histo-pathology for Widely Generalizable Model}

\author{
  Jevgenij Gamper\\
  Mathematics of Systems\\
  Warwick University\\
  Coventry, UK \\
  \texttt{j.gamper@warwick.ac.uk} \\
   \And
 Navid Alemi Kooohbanani \quad Nasir Rajpoot \\
  Department of Computer Science\\
  Warwick University\\
  Coventry, UK \\
}

\begin{document}

\maketitle

\begin{abstract}
In this work we show preliminary results of deep multi-task learning in the area of computational pathology. We combine 11 tasks ranging from patch-wise oral cancer classification, one of the most prevalent cancers in the developing world, to multi-tissue nuclei instance segmentation and classification. 
\end{abstract}

\section{Introduction}

The emerging area of computational pathology (CPath) is ripe ground for the application of deep learning methods to healthcare due to the sheer volume of raw pixel data in whole-slide images (WSIs) of cancerous tissue slides, generally of the order of $100K{\times}80K$ pixels \citep{colling2019artificial}. However, despite the availability of raw pixel data in CPath, the ground truth for training AI models is sparse, expensive to obtain, and noisy. Diverse multi-centre data are usually limited to the developed world; yet, developing countries are where AI could find their  most viable application amid global shortages of clinical experts \citep{coelho2012challenges} and widespread incidence of oral cancer \citep{shrivastava2014oral} and lymphomas \citep{perry2016non}.

A major challenge facing wider adoption of AI, especially in a resource-strapped setting and in the absence of well-curated multi-centric high-quality datasets, is algorithm robustness due to the lack of feature generalisation \citep{zech2018variable}. Some recent studies have pointed to limited clinical applicability of AI due to weak experimental design even with datasets obtained in the developed world \citep{liu2019comparison, zech2018variable}. Additionally, AI research points to several vulnerabilities of deep learning models widely adopted in the aforementioned studies \citep{geirhos2018imagenet, ghorbani2019interpretation}. As such, particularly for the application within the developing countries, AI models must be robust and learn semantically meaningful features, in order to be able to generalise across a variety of tasks. It is thus reasonable to maximise the utility of available good quality datasets in the literature in order to test the feasibility of obtaining such a model. We propose to test deep multi-task learning (MTL) as a method to obtain general feature representation that would be applicable to new tasks or tasks with relatively small amounts of data \citep{raghu2019transfusion}.  We present preliminary results of a simple, yet promising approach to MTL in CPath and has potential to do well particularly on oral cancer, one of the most important cancers of relevance to the developing world. 

\begin{table}[!h]
\centering
\caption{Pixel-wise segmentation and image classification datasets, their sources, baseline results obtained from the literature and un-tuned MTL approach results.}
\label{table:baseline_vs_mtl}
\begin{tabular}{ccccc} 
Name (Source) & Task                                            & Baseline & MTL  \\ 
\hline
              &                                                 &          &          \\
              & Pixel-level Tasks                               & \multicolumn{3}{c}{PQ}      \\ 
\hline
\cite{awan2017glandular}          & Gland Segmentation                              & 0.76    & 0.60     \\
\cite{gamper2019pannuke}         & Nuclei Inst. Segm. and Class. & 0.38    & 0.18      \\
\cite{janowczyk2016deep}              & Epithelium Segmentation                         & 0.68    & 0.57      \\
\cite{fraz2019fabnet}              & Vessel Segmentation                             & 0.67    & 0.59       \\ 
\hline
              &                                                 &          &        &           \\
              & Patch-level Classication Tasks                  & \multicolumn{3}{c}{Accuracy}  \\ 
\hline
\cite{litjens20181399}     & Breast (2)                                      & 92.44   & 91.23       \\
\cite{kather2019predicting}              & Colon (9)                                       & 98.70    & 89.00       \\
\cite{alsubaie2018multi}              & Lung (6)                                        & 91.00    & 78.56     \\
\cite{janowczyk2016deep}              & Lymphoma (3)                                    & 96.58    & 58.22      \\
\cite{qureshi2008adaptive}              & Meningioma (4)                                  & 82.10    & 92.19       \\
\cite{shaban2019novel}              & Oral (3)                                        & 96.30    & 91.11     \\
\cite{kobel2010diagnosis}              & Ovary (5)                                       & 89.40    & 72.56     
\end{tabular}
\end{table}

\section{Materials \& Methods}
\citet{baxter2000model} theoretically showed that learning from multiple related tasks results in fast learning as measured by the number of training examples required per task; and that inductive biases learned on sufficiently many training tasks will likely generalise to novel tasks. \citet{subramanian2018learning} demonstrated successful application of the aforementioned theory to natural language processing via deep MTL. \citet{brenes2019histomulty} demonstrated the potential of MTL in histology to obtain robust features using only two tasks. In this work, we collect 11 well-established tasks ranging from image classification to pixel-wise instance segmentation and classification (see Table \ref{table:baseline_vs_mtl}). 

\begin{algorithm}[h]
\label{algo}
\KwIn{A set of $k$ histology tasks, their corresponding datasets $\mathbb{P}_1, ..., \mathbb{P}_k$ and a set of $k$ task specific decoders $\mathbf{D}_1 . . . \mathbf{D}_k$, a feature encoder $\mathbf{E}$ shared across all tasks. Let $\theta$ denote model parameters (encoder and decoders), and $\alpha$ be a probability vector $(p_1, ..., p_k)$ denoting the probability of sampling a task at a given iteration such that $\sum_{i=1}^{k} p_i = 1$. Let $L$ and $T$ denote the loss function and a maximum number of iterations, respectively.}
\KwOut{Trained encoder $\mathbf{E}$ and decoders $\mathbf{D}_1 . . . \mathbf{D}_k$. }
\Repeat{$t$ = $T$}{$t+=1$ \\
Sample a task $i \sim \text{Cat}(k, \alpha)$ \\
Sample input, output pairs $\mathbf{x}, \mathbf{y} \sim \mathbb{P}_i$ \\
Encode inputs $h_\mathbf{x} \leftarrow \mathbf{E}_{\theta}(\mathbf{x)}$ \\ 
Predict $\hat{\mathbf{y}} \leftarrow \mathbf{D}_{i,\theta}(\mathbf{h}_{\mathbf{x}})$ \\
Update $\theta \leftarrow \text{Adam}(\nabla_{\theta} L(\mathbf{y} , \hat{\mathbf{y}}))$}
    \caption{{\bf Multi-Task Training Setup} \label{Algorithm}}
\end{algorithm}

To quantify the performance of segmentation tasks we use panoptic quality (PQ) \cite{kirillov2019panoptic}, and classification accuracy for image classification tasks. Baseline results are single-task results as quoted in literature. Epithelium segmentation PQ however was not available, we thus obtained it by training a single task segmentation model. All segmentation decoders were based on Pyramid Scene Parsing network \citep{zhao2017pyramid}, and classification decoders were simply a fully connected layer followed by softmax or sigmoid non-linearity. Each decoder processed 2048 dimensional features extracted using a trainable Resnet-50 encoder \citep{he2016deep}. Our MTL optimisation approach is described formally in the Algorithm \ref{algo} above. In summary, at every optimisation step we randomly pick task index, and use the corresponding task dataset to extract data, process input using the encoder and decode it using task specific decoder, followed by the evaluation of the loss function and gradient update using Adam \citep{kingma2014adam}.

\section{Results \& Conclusion}

\begin{figure}[h]
\begin{center}
   \includegraphics[width=0.8\linewidth]{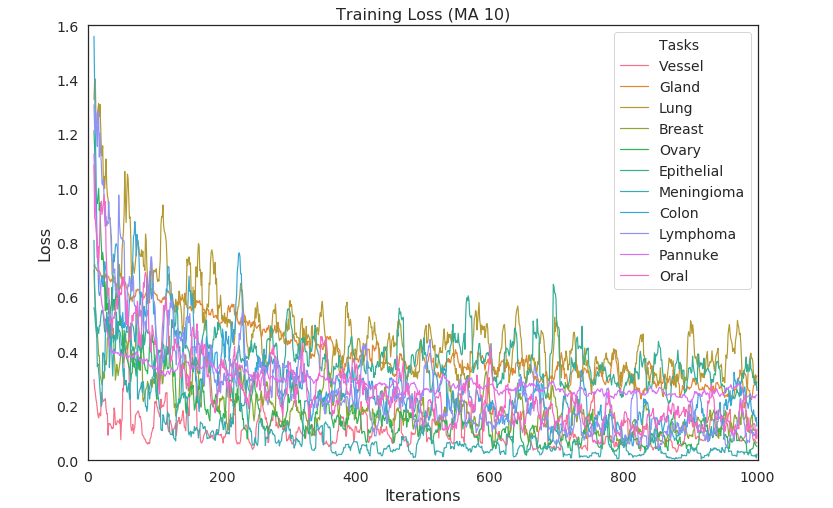}
\end{center}
   \caption{Loss over the first 1000 iterations, due to high variance of the loss it has been smoothed using rolling average with window of size 10.}
\label{fig:loss}
\end{figure}

\begin{figure}[h]
\begin{center}
   \includegraphics[width=0.8\linewidth]{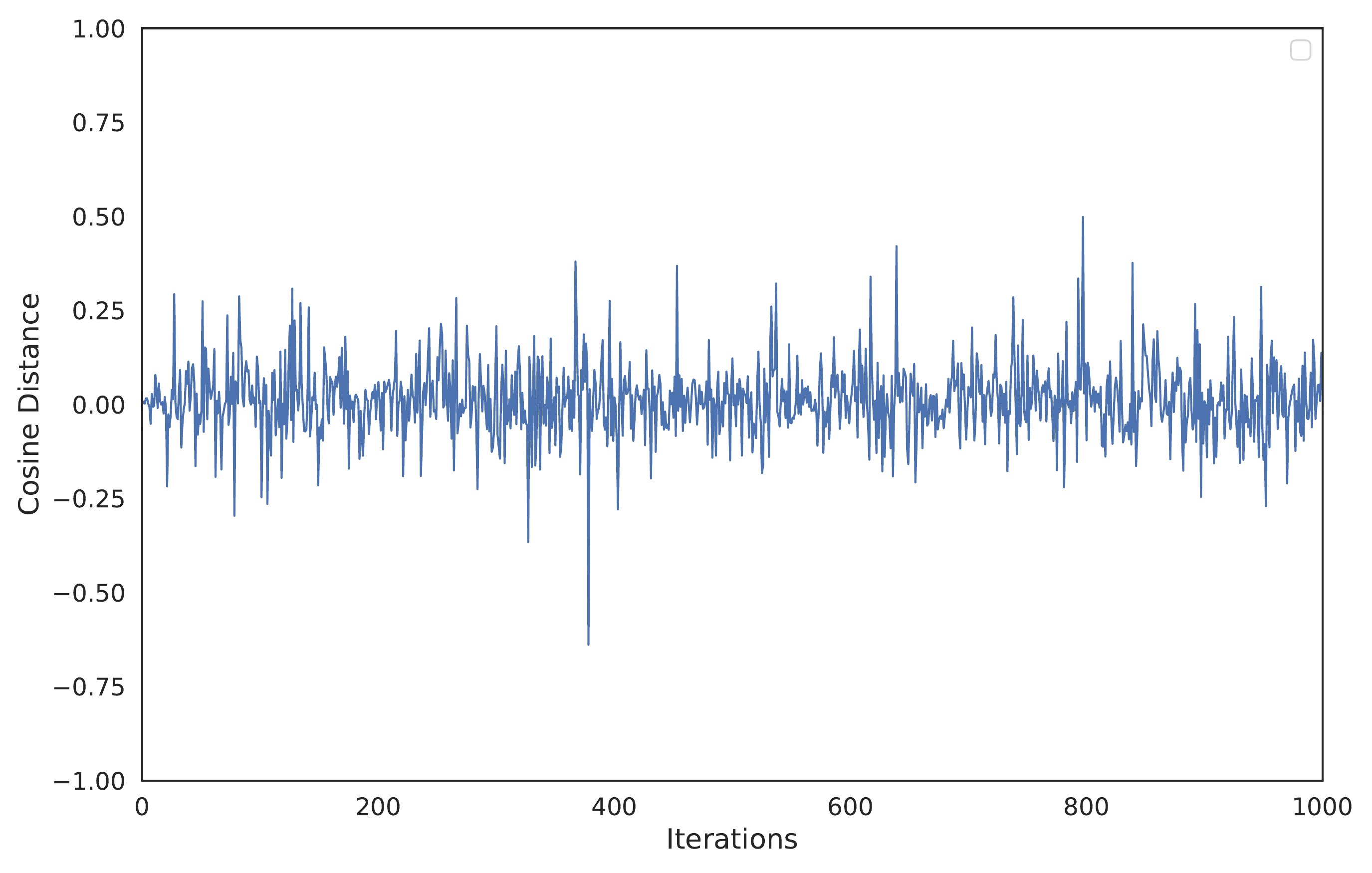}
\end{center}
   \caption{Cosine distance between encoder gradients for task sampled at iteration $t$ and task at iteration $t-1$.}
\label{fig:cosine-iterations}
\end{figure}

\begin{figure}[h]
\begin{center}
   \includegraphics[width=1\linewidth]{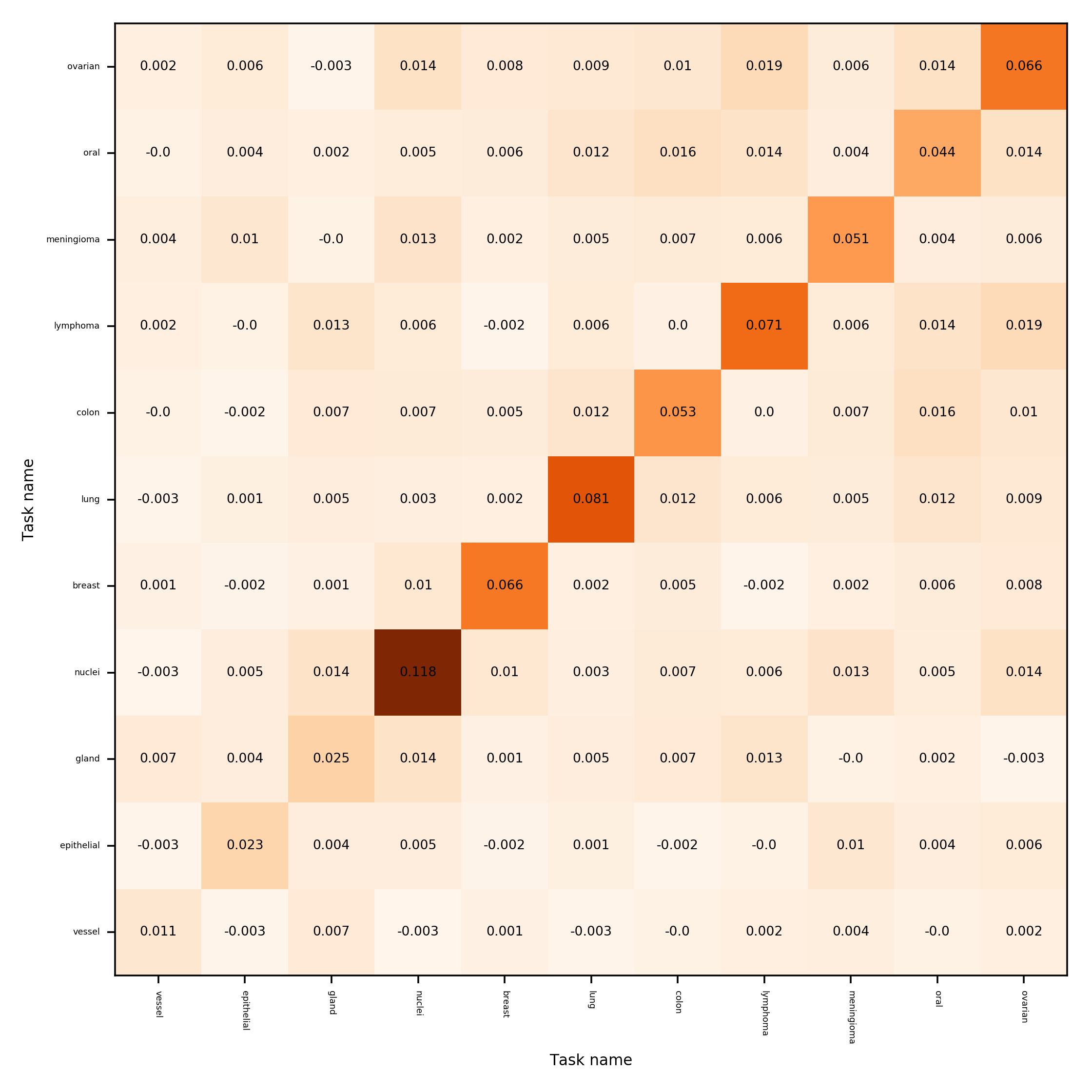}
\end{center}
   \caption{Rolling mean cosine distance between task gradients of the encoder.}
\label{fig:cosine-matrix}
\end{figure}

 Our preliminary results of deep multi-task training with default hyper-parameters and no-tuning are presented in Table \ref{table:baseline_vs_mtl} under column MTL. Results are encouraging, particularly given the variance of the loss during training as presented in Figure A\ref{fig:loss}. Loss variance has been attributed to the direction of gradients and has been widely studied in lifelong learning and MTL \citep{lopez2017gradient, du2018adapting, chaudhry2018efficient, yu2020gradient}. Our preliminary results in Figure A\ref{fig:cosine-matrix} and Figure A\ref{fig:cosine-iterations} demonstrate small cosine distances. However, distribution of distances between gradient vectors becomes narrowly focused around zero as the dimensionality grows. The significance of small distances may increase with growing model size. In the future work we will further investigate MTL optimisation characteristics and match single task performance, as well as test the generalisation of obtained features.

\bibliography{iclr2020_conference}

\begin{thebibliography}{30}
\providecommand{\natexlab}[1]{#1}
\providecommand{\url}[1]{\texttt{#1}}
\expandafter\ifx\csname urlstyle\endcsname\relax
  \providecommand{\doi}[1]{doi: #1}\else
  \providecommand{\doi}{doi: \begingroup \urlstyle{rm}\Url}\fi

\bibitem[Alsubaie et~al.(2018)Alsubaie, Shaban, Snead, Khurram, and
  Rajpoot]{alsubaie2018multi}
Najah Alsubaie, Muhammad Shaban, David Snead, Ali Khurram, and Nasir Rajpoot.
\newblock A multi-resolution deep learning framework for lung adenocarcinoma
  growth pattern classification.
\newblock In \emph{Annual Conference on Medical Image Understanding and
  Analysis}, pp.\  3--11. Springer, 2018.

\bibitem[Awan et~al.(2017)Awan, Sirinukunwattana, Epstein, Jefferyes, Qidwai,
  Aftab, Mujeeb, Snead, and Rajpoot]{awan2017glandular}
Ruqayya Awan, Korsuk Sirinukunwattana, David Epstein, Samuel Jefferyes, Uvais
  Qidwai, Zia Aftab, Imaad Mujeeb, David Snead, and Nasir Rajpoot.
\newblock Glandular morphometrics for objective grading of colorectal
  adenocarcinoma histology images.
\newblock \emph{Scientific reports}, 7\penalty0 (1):\penalty0 1--12, 2017.

\bibitem[Baxter(2000)]{baxter2000model}
Jonathan Baxter.
\newblock A model of inductive bias learning.
\newblock \emph{Journal of artificial intelligence research}, 12:\penalty0
  149--198, 2000.

\bibitem[Brenes(2019)]{brenes2019histomulty}
David Brenes.
\newblock Multi-task deep learning model for improved histopathology prediction
  from in-vivo microscopy images.
\newblock \emph{LXAI workshop at NeurIPS}, 2019.

\bibitem[Chaudhry et~al.(2018)Chaudhry, Ranzato, Rohrbach, and
  Elhoseiny]{chaudhry2018efficient}
Arslan Chaudhry, Marc'Aurelio Ranzato, Marcus Rohrbach, and Mohamed Elhoseiny.
\newblock Efficient lifelong learning with a-gem.
\newblock \emph{arXiv preprint arXiv:1812.00420}, 2018.

\bibitem[Coelho(2012)]{coelho2012challenges}
Ken~Russell Coelho.
\newblock Challenges of the oral cancer burden in india.
\newblock \emph{Journal of cancer epidemiology}, 2012, 2012.

\bibitem[Colling et~al.(2019)Colling, Pitman, Oien, Rajpoot, Macklin, Snead,
  Sackville, Verrill, in~Histopathology Working~Group, Bachtiar,
  et~al.]{colling2019artificial}
Richard Colling, Helen Pitman, Karin Oien, Nasir Rajpoot, Philip Macklin, David
  Snead, Tony Sackville, Clare Verrill, CM-Path~AI in~Histopathology
  Working~Group, Velicia Bachtiar, et~al.
\newblock Artificial intelligence in digital pathology: A roadmap to routine
  use in clinical practice.
\newblock \emph{The Journal of pathology}, 2019.

\bibitem[Du et~al.(2018)Du, Czarnecki, Jayakumar, Pascanu, and
  Lakshminarayanan]{du2018adapting}
Yunshu Du, Wojciech~M Czarnecki, Siddhant~M Jayakumar, Razvan Pascanu, and
  Balaji Lakshminarayanan.
\newblock Adapting auxiliary losses using gradient similarity.
\newblock \emph{arXiv preprint arXiv:1812.02224}, 2018.

\bibitem[Fraz et~al.(2019)Fraz, Khurram, Graham, Shaban, Hassan, Loya, and
  Rajpoot]{fraz2019fabnet}
MM~Fraz, SA~Khurram, S~Graham, M~Shaban, M~Hassan, A~Loya, and NM~Rajpoot.
\newblock Fabnet: feature attention-based network for simultaneous segmentation
  of microvessels and nerves in routine histology images of oral cancer.
\newblock \emph{Neural Computing and Applications}, pp.\  1--14, 2019.

\bibitem[Gamper et~al.(2019)Gamper, Koohbanani, Benet, Khuram, and
  Rajpoot]{gamper2019pannuke}
Jevgenij Gamper, Navid~Alemi Koohbanani, Ksenija Benet, Ali Khuram, and Nasir
  Rajpoot.
\newblock Pannuke: an open pan-cancer histology dataset for nuclei instance
  segmentation and classification.
\newblock In \emph{European Congress on Digital Pathology}, pp.\  11--19.
  Springer, 2019.

\bibitem[Geirhos et~al.(2018)Geirhos, Rubisch, Michaelis, Bethge, Wichmann, and
  Brendel]{geirhos2018imagenet}
Robert Geirhos, Patricia Rubisch, Claudio Michaelis, Matthias Bethge, Felix~A
  Wichmann, and Wieland Brendel.
\newblock Imagenet-trained cnns are biased towards texture; increasing shape
  bias improves accuracy and robustness.
\newblock \emph{arXiv preprint arXiv:1811.12231}, 2018.

\bibitem[Ghorbani et~al.(2019)Ghorbani, Abid, and
  Zou]{ghorbani2019interpretation}
Amirata Ghorbani, Abubakar Abid, and James Zou.
\newblock Interpretation of neural networks is fragile.
\newblock In \emph{Proceedings of the AAAI Conference on Artificial
  Intelligence}, volume~33, pp.\  3681--3688, 2019.

\bibitem[He et~al.(2016)He, Zhang, Ren, and Sun]{he2016deep}
Kaiming He, Xiangyu Zhang, Shaoqing Ren, and Jian Sun.
\newblock Deep residual learning for image recognition.
\newblock In \emph{Proceedings of the IEEE conference on computer vision and
  pattern recognition}, pp.\  770--778, 2016.

\bibitem[Janowczyk \& Madabhushi(2016)Janowczyk and
  Madabhushi]{janowczyk2016deep}
Andrew Janowczyk and Anant Madabhushi.
\newblock Deep learning for digital pathology image analysis: A comprehensive
  tutorial with selected use cases.
\newblock \emph{Journal of pathology informatics}, 7, 2016.

\bibitem[Kather et~al.(2019)Kather, Krisam, Charoentong, Luedde, Herpel, Weis,
  Gaiser, Marx, Valous, Ferber, et~al.]{kather2019predicting}
Jakob~Nikolas Kather, Johannes Krisam, Pornpimol Charoentong, Tom Luedde,
  Esther Herpel, Cleo-Aron Weis, Timo Gaiser, Alexander Marx, Nektarios~A
  Valous, Dyke Ferber, et~al.
\newblock Predicting survival from colorectal cancer histology slides using
  deep learning: A retrospective multicenter study.
\newblock \emph{PLoS medicine}, 16\penalty0 (1), 2019.

\bibitem[Kingma \& Ba(2014)Kingma and Ba]{kingma2014adam}
Diederik~P Kingma and Jimmy Ba.
\newblock Adam: A method for stochastic optimization.
\newblock \emph{arXiv preprint arXiv:1412.6980}, 2014.

\bibitem[Kirillov et~al.(2019)Kirillov, He, Girshick, Rother, and
  Doll{\'a}r]{kirillov2019panoptic}
Alexander Kirillov, Kaiming He, Ross Girshick, Carsten Rother, and Piotr
  Doll{\'a}r.
\newblock Panoptic segmentation.
\newblock In \emph{Proceedings of the IEEE Conference on Computer Vision and
  Pattern Recognition}, pp.\  9404--9413, 2019.

\bibitem[K{\"o}bel et~al.(2010)K{\"o}bel, Kalloger, Baker, Ewanowich, Arseneau,
  Zherebitskiy, Abdulkarim, Leung, Duggan, Fontaine,
  et~al.]{kobel2010diagnosis}
Martin K{\"o}bel, Steve~E Kalloger, Patricia~M Baker, Carol~A Ewanowich,
  Jocelyne Arseneau, Viktor Zherebitskiy, Soran Abdulkarim, Samuel Leung,
  M{\'a}ire~A Duggan, Dan Fontaine, et~al.
\newblock Diagnosis of ovarian carcinoma cell type is highly reproducible: a
  transcanadian study.
\newblock \emph{The American journal of surgical pathology}, 34\penalty0
  (7):\penalty0 984--993, 2010.

\bibitem[Litjens et~al.(2018)Litjens, Bandi, Ehteshami~Bejnordi, Geessink,
  Balkenhol, Bult, Halilovic, Hermsen, van~de Loo, Vogels,
  et~al.]{litjens20181399}
Geert Litjens, Peter Bandi, Babak Ehteshami~Bejnordi, Oscar Geessink, Maschenka
  Balkenhol, Peter Bult, Altuna Halilovic, Meyke Hermsen, Rob van~de Loo, Rob
  Vogels, et~al.
\newblock 1399 h\&e-stained sentinel lymph node sections of breast cancer
  patients: the camelyon dataset.
\newblock \emph{GigaScience}, 7\penalty0 (6):\penalty0 giy065, 2018.

\bibitem[Liu et~al.(2019)Liu, Faes, Kale, Wagner, Fu, Bruynseels, Mahendiran,
  Moraes, Shamdas, Kern, et~al.]{liu2019comparison}
Xiaoxuan Liu, Livia Faes, Aditya~U Kale, Siegfried~K Wagner, Dun~Jack Fu, Alice
  Bruynseels, Thushika Mahendiran, Gabriella Moraes, Mohith Shamdas, Christoph
  Kern, et~al.
\newblock A comparison of deep learning performance against health-care
  professionals in detecting diseases from medical imaging: a systematic review
  and meta-analysis.
\newblock \emph{The Lancet Digital Health}, 1\penalty0 (6):\penalty0
  e271--e297, 2019.

\bibitem[Lopez-Paz \& Ranzato(2017)Lopez-Paz and Ranzato]{lopez2017gradient}
David Lopez-Paz and Marc'Aurelio Ranzato.
\newblock Gradient episodic memory for continual learning.
\newblock In \emph{Advances in Neural Information Processing Systems}, pp.\
  6467--6476, 2017.

\bibitem[Perry et~al.(2016)Perry, Diebold, Nathwani, MacLennan,
  M{\"u}ller-Hermelink, Bast, Boilesen, Armitage, and
  Weisenburger]{perry2016non}
Anamarija~M Perry, Jacques Diebold, Bharat~N Nathwani, Kenneth~A MacLennan,
  Hans~K M{\"u}ller-Hermelink, Martin Bast, Eugene Boilesen, James~O Armitage,
  and Dennis~D Weisenburger.
\newblock Non-hodgkin lymphoma in the developing world: review of 4539 cases
  from the international non-hodgkin lymphoma classification project.
\newblock \emph{Haematologica}, 101\penalty0 (10):\penalty0 1244--1250, 2016.

\bibitem[Qureshi et~al.(2008)Qureshi, Sertel, Rajpoot, Wilson, and
  Gurcan]{qureshi2008adaptive}
Hammad Qureshi, Olcay Sertel, Nasir Rajpoot, Roland Wilson, and Metin Gurcan.
\newblock Adaptive discriminant wavelet packet transform and local binary
  patterns for meningioma subtype classification.
\newblock In \emph{International Conference on Medical Image Computing and
  Computer-Assisted Intervention}, pp.\  196--204. Springer, 2008.

\bibitem[Raghu et~al.(2019)Raghu, Zhang, Kleinberg, and
  Bengio]{raghu2019transfusion}
Maithra Raghu, Chiyuan Zhang, Jon Kleinberg, and Samy Bengio.
\newblock Transfusion: Understanding transfer learning for medical imaging.
\newblock In \emph{Advances in Neural Information Processing Systems}, pp.\
  3342--3352, 2019.

\bibitem[Shaban et~al.(2019)Shaban, Khurram, Fraz, Alsubaie, Masood, Mushtaq,
  Hassan, Loya, and Rajpoot]{shaban2019novel}
Muhammad Shaban, Syed~Ali Khurram, Muhammad~Moazam Fraz, Najah Alsubaie, Iqra
  Masood, Sajid Mushtaq, Mariam Hassan, Asif Loya, and Nasir~M Rajpoot.
\newblock A novel digital score for abundance of tumour infiltrating
  lymphocytes predicts disease free survival in oral squamous cell carcinoma.
\newblock \emph{Scientific reports}, 9\penalty0 (1):\penalty0 1--13, 2019.

\bibitem[Shrivastava et~al.(2014)Shrivastava, Shrivastava, and
  Ramasamy]{shrivastava2014oral}
Saurabh~R Shrivastava, Prateek~S Shrivastava, and Jegadeesh Ramasamy.
\newblock Oral cancer in developing countries: the time to act is upon us.
\newblock \emph{Iranian journal of cancer prevention}, 7\penalty0 (1):\penalty0
  58, 2014.

\bibitem[Subramanian et~al.(2018)Subramanian, Trischler, Bengio, and
  Pal]{subramanian2018learning}
Sandeep Subramanian, Adam Trischler, Yoshua Bengio, and Christopher~J Pal.
\newblock Learning general purpose distributed sentence representations via
  large scale multi-task learning.
\newblock \emph{arXiv preprint arXiv:1804.00079}, 2018.

\bibitem[Yu et~al.(2020)Yu, Kumar, Gupta, Levine, Hausman, and
  Finn]{yu2020gradient}
Tianhe Yu, Saurabh Kumar, Abhishek Gupta, Sergey Levine, Karol Hausman, and
  Chelsea Finn.
\newblock Gradient surgery for multi-task learning.
\newblock \emph{arXiv preprint arXiv:2001.06782}, 2020.

\bibitem[Zech et~al.(2018)Zech, Badgeley, Liu, Costa, Titano, and
  Oermann]{zech2018variable}
John~R Zech, Marcus~A Badgeley, Manway Liu, Anthony~B Costa, Joseph~J Titano,
  and Eric~Karl Oermann.
\newblock Variable generalization performance of a deep learning model to
  detect pneumonia in chest radiographs: a cross-sectional study.
\newblock \emph{PLoS medicine}, 15\penalty0 (11), 2018.

\bibitem[Zhao et~al.(2017)Zhao, Shi, Qi, Wang, and Jia]{zhao2017pyramid}
Hengshuang Zhao, Jianping Shi, Xiaojuan Qi, Xiaogang Wang, and Jiaya Jia.
\newblock Pyramid scene parsing network.
\newblock In \emph{Proceedings of the IEEE conference on computer vision and
  pattern recognition}, pp.\  2881--2890, 2017.

\end{thebibliography}
\bibliographystyle{iclr2020_conference}

\newpage

\appendix
\section{Appendix}

\end{document}